\documentclass[runningheads]{llncs}
\usepackage[T1]{fontenc}
\usepackage{graphicx}

\usepackage[colorlinks,urlcolor=blue,linkcolor=blue,citecolor=blue]{hyperref}

\usepackage{booktabs}       
\usepackage{subcaption}

\usepackage{ifthen}  

\usepackage{xr}
\graphicspath{ {./figures/seed_runs/} }

\def\citep{\cite}
\def\citet{\cite}

\begin{document}

\title{Graceful task adaptation with a bi-hemispheric RL agent}

%

\newboolean{anon}
\setboolean{anon}{false} 

\ifthenelse{\boolean{anon}}
{
\author{First Author\inst{1}\orcidID{0000-1111-2222-3333} \and
Second Author\inst{2,3}\orcidID{1111-2222-3333-4444} \and
Third Author\inst{3}\orcidID{2222--3333-4444-5555}}
\authorrunning{F. Author et al.}
%
\institute{Institute 1
\email{email@email.com} \and
Institute 2 
\email{\{abc,lncs\}@email.com}
\url{http://www.institute.com} }
}
{

\author{
  Grant Nicholas\inst{1}
  \and
  Levin Kuhlmann\inst{1}\orcidID{0000-0002-5108-6348}
  \and
  Gideon Kowadlo\inst{1,2}\orcidID{0000-0001-6036-1180}
}

\authorrunning{G. Nicholas et al.}

\institute{Monash University
\email{grant.d.nicholas@gmail.com} \\
\email{levin.kuhlmann@monash.edu} \\
\and
Cerenaut
\email{gideon@cerenaut.ai} 
\url{https://cerenaut.ai}
}
}

\maketitle

\begin{abstract}
In humans, responsibility for performing a task gradually shifts from the right hemisphere to the left. The Novelty-Routine Hypothesis (NRH) states that the right and left hemispheres are used to perform novel and routine tasks respectively, enabling us to learn a diverse range of novel tasks while performing the task capably. Drawing on the NRH, we develop a reinforcement learning agent with specialised hemispheres that can exploit generalist knowledge from the right-hemisphere to avoid poor initial performance on novel tasks. In addition, we find that this design has minimal impact on its ability to learn novel tasks. We conclude by identifying improvements to our agent and exploring potential expansion to the continual learning setting.
\end{abstract}

\section{Introduction}
Despite many high-profile successes of Deep Reinforcement Learning (RL), RL-agents still struggle with sample efficiency and generalisation \cite{botvinick_reinforcement_2019}. Human beings do not experience such problems and are capable continuous learners, able to acquire diverse skills over their lifetimes \cite{khetarpal_towards_2022}. 

Research in neuroscience highlights the importance of hemispheric specialisation in the human brain for learning \cite{goldberg_hemisphere_1981,goldberg_lateralization_1994,gazzaniga_cerebral_2000,monaghan_hemispheric_2004}. 
The right-hemisphere is identified as focusing on `global' phenomena, being competent in novel scenarios and for exploration, while the left-hemisphere focuses on `local' phenomena and learns to specialise \cite{goldberg_hemisphere_1981,goldberg_lateralization_1994,gazzaniga_cerebral_2000,peleg_two_2010}. 
When a novel task is learnt, a uni-directional shift occurs whereby responsibility for the task moves from right to left hemisphere during learning \cite{goldberg_lateralization_1994}; the key to the Novelty Routine Hypothesis (NRH) proposed in \citet{goldberg_lateralization_1994}.

Consequently, we draw on the NRH and related neuroscientific theories to develop an RL-agent with specialised hemispheres. In doing so, we follow the approach of other significant discoveries in AI that use human neuroscience as inspiration for AI algorithm development \cite{parisi_continual_2019,hassabis_neuroscience-inspired_2017}. We hypothesise that our bi-hemispheric architecture will help RL-agents learn novel tasks, while avoiding the poor initial performance typically observed for agents trained from scratch. Essentially, bi-hemispheric agents should perform well in the initial stages of learning by drawing on the generalist right-hemisphere, while training the left-hemisphere to specialise in the task. This may have benefits for continual learning, where agents encounter streams of novel tasks.

Little research in RL explores the application of bi-hemispheric brain structure. Mixture of Experts (MoE)
\cite{jacobs_adaptive_1991} and other multiple network architectures are neuroscience inspired, but do not focus on hemispheric specialisation \cite{pham_dualnet_2021,tsuda_modeling_2020}. Furthermore, while Actor-Critic based approaches can use dual networks to output actions and values, the aim is to stabilise estimates of the policy gradient, rather than mimic brain structure \cite{mnih_asynchronous_2016}. In supervised learning, Rajagopalan et al. \citet{rajagopalan_deep_2022} explored a bi-hemispheric ensemble model with specialist left and generalist right hemispheres to classify images. Neuroscientists also use bi-hemispheric deep neural networks to test hypotheses about the human brain \cite{shevtsova_neural_1999,weems_hemispheric_2004,peleg_two_2010,chang_unified_2020}. This project is a novel application of a bi-hemispheric architecture to RL.

\section{Agent design} 
\label{sec:approach}
We constructed a bi-hemispheric agent inspired by the NRH; a right-hemisphere has generalist capabilities and a left-hemisphere learns to specialise in tasks. When encountering a novel task, a gating network assigns responsibility to either the right or left hemisphere to manage performance. The agent should draw on generalist skills in the right-hemisphere to achieve improved initial performance over an agent trained from scratch. This should not interfere with using the left-hemisphere to learn and eventually perform the novel task independently. Hence, the agent is assessed on the \textbf{objectives}:

\begin{enumerate}
    \item \emph{initial} performance better than an agent trained from scratch
    \item \emph{final} left-hemisphere performance as good as an agent trained from scratch
\end{enumerate}

\subsection{Network architecture}
The bi-hemispheric agent consists of two hemispheres and a gating network, each comprising a Recurrent Neural Network (RNN), with Gated Recurrent Units (GRU). Each hemisphere outputs an action and a value estimate. For simplicity, we give right and left hemispheres identical architectures and omit inter-hemispheric connections (i.e. Corpus Callosum), despite being neurologically inaccurate \cite{goldberg_hemispheric_2013,chang_unified_2020,peleg_two_2010,gazzaniga_cerebral_2000}. 
The network architecture is shown in Figure~S.\ref{fig:network_arch}.

\subsection{Gating network} \label{sec:gating_func}
The gating network assigns `responsibility' to each hemisphere using `gating values', defined as the proportional contribution of that hemisphere to the agent's action and value. The responsibility of the right-hemisphere is thus:
\begin{equation}
    P^{right} = 1 - P^{left}\;where\;P^{right}, P^{left} \in [0,1] 
\end{equation}

The value estimate is a linear combination of the right and left hemisphere value outputs, weighted by the responsibility of each hemisphere:
\begin{equation}
    V^{bihem} = P^{right}V^{right} + P^{left}V^{left}   
\end{equation}

Similarly, our bi-hemipheric agent's policy samples actions from a Gaussian distribution whose mean is also a linear combination of right and left hemisphere means. The variance is \(\sigma^{2}\) learnt by the bi-hemispheric agent.
\begin{equation}
    \pi^{bihem} \sim \mathcal{N}(P^{right}\mu^{right} + P^{left}\mu^{left}, \sigma^{2})
\end{equation}

The gating network takes as input the gating values from the last timestep and the `value-estimate error' \(\varepsilon\) for each hemisphere \(h\), defined as \(\varepsilon^{h} = V^{h}_{t} - r_{t}\). This gives the gating network information on how well the hemisphere's can predict task rewards. Combining this with previous gating values lets the gating network learn relationships between hemisphere performance and responsibility.

Finally, to encourage hemispheric shift we incorporate an additive penalty into the bi-hemispheric agent's loss function. This term penalises losses when the right gating value is large:
\begin{equation}
   \beta(\frac{P^{right}}{P^{left}})^{\alpha}
\end{equation}
Where \(\alpha\) and \(\beta\) are tuneable parameters. For details see Section~S.\ref{supp:loss}.

\subsection{Generalisation and specialisation}
\label{sec:gen_spec}
We trained right and left hemispheres separately using different training processes to induce generalisation and specialisation respectively. 
First, we trained the right-hemisphere using the \(RL^{2}\) meta-learning algorithm \cite{duan_rl2_2016,wang_learning_2016}. 
Second, we constructed the bi-hemispheric network by freezing the weights of the right-hemisphere and then combined it with a randomly initialised left-hemisphere and gating network.
Finally, the left-hemisphere and gating networks were trained together using the standard RL objective, which maximises expected discounted rewards \cite{sutton_reinforcement_2018}.
The left-hemisphere `specialises' in a task simply by being trained to perform that specific task.

We used meta-learning for the right, as it offers both the ability to generalise to a task-distribution \emph{and} adapt quickly to novel tasks \cite{finn_model-agnostic_2017}; both important for strong initial performance. \(RL^{2}\) is exceptional at the latter -- sometimes achieving zero-shot adaptation \cite{duan_rl2_2016,wang_learning_2016,kirk_survey_2023}. In addition \(RL^{2}\) is a memory-based meta-RL algorithm, which have been identified as Bayesian-optimal learners as they can optimally trade off exploration and exploitation in uncertain environments \cite{zintgraf_varibad_2021,ortega_meta-learning_2019}. This fits with the NRH's view of the right-hemisphere as guiding exploratory behaviour when learning novel tasks \cite{goldberg_lateralization_1994}.

\section{Experiments}
\label{sec:methodology}
\begin{figure}
  \centering
  \includegraphics[width=0.8\textwidth]{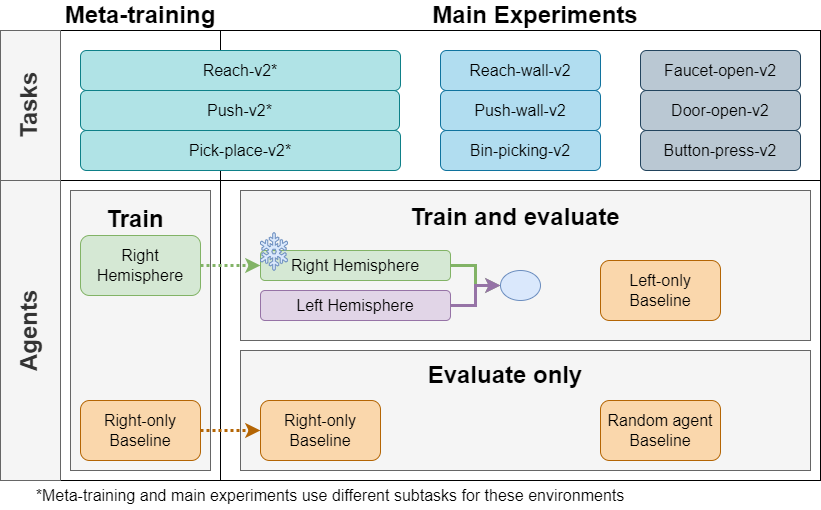}
  \caption{\small Experiment overview}
  \label{fig:experiment_approach}
\end{figure}

Our implementation is available at 
\ifthenelse{\boolean{anon}}
{
\url{https://anonymous.4open.science/r/right_left_brain_rl-328E}.
}
{
\url{https://github.com/gdubbs100/right_left_brain_rl}
}
The experiments were carried out in two stages, shown in Figure~\ref{fig:experiment_approach}. We first meta-trained the right-hemisphere and a baseline agent on 3 tasks. We then created the bi-hemispheric agent and trained and evaluated all agents on an expanded set of tasks over 5 seeds.

\subsection{Dataset} 
\label{sec:exp_envs}

We used Meta-world \cite{yu_meta-world_2020}, a meta-learning/multi-task benchmark comprised of 50 tasks that involve manipulation of objects using a Sawyer robot arm. We chose Meta-world because of its diverse tasks with shared action and state spaces. 

Meta-world tasks are designed to vary in a non-trivial manner, to allow the evaluation of generalisation capabilities in RL-agents. Meta-world defines differences between tasks as parametric or non-parametric. Parametric variation involves variation in real-valued parameters such as object or goal locations, e.g. the location of a ball where the task objective is for an arm to reach said ball. In contrast, non-parametric variation involves qualitative differences between tasks, e.g. the difference between opening a door and lifting a ball \cite{yu_meta-world_2020}. Each of Meta-world's 50 tasks exhibit non-parametric variation between each other. Further, each task is comprised of 50 sub-tasks which exhibit parametric variation between each other. 
We use Meta-world's definitions of variation to divide the experiment tasks into three tiers (Section~S.\ref{supp:tasks}) and classify the novelty of a task relative to the meta-training tasks: 

\begin{enumerate}
    \item \textbf{Tier-1} tasks have only parametric variation i.e. sub-tasks with different goal and object locations. 
    \item \textbf{Tier-2} tasks include similar tasks to Tier-1, but with minor non-parametric variation e.g. the addition of a wall obstacle to a task from meta-training.
    \item \textbf{Tier-3} tasks exhibit complete non-parametric variation from Tier-1. These tasks help us understand the limits of our approach.
\end{enumerate}

\subsection{Baselines}
We evaluated three baselines. First, the `left-only' baseline is a randomly initialised agent with the same architecture as the left-hemisphere of the bi-hemispheric agent, but double the size of the GRU (i.e. increased neurons to 256 from 128). We double GRU size to ensure the same size network as the bi-hemispheric agents, giving them equal representational capacity and enabling a fair comparison between agents. We also evaluated a `right-only' baseline -- a meta-trained agent with identical architecture and doubled GRU size. This baseline let us determine whether it is better to simply meta-train an agent rather than use the bi-hemispheric architecture. Finally, we used a randomly acting agent as a lower-bound on performance; details in Section~S.\ref{supp:agents}.

\subsection{Training approach}

\subsubsection{Meta-training} 
\label{sec:meta_train_meth}
We meta-trained the right-hemisphere and right-only baseline using the \(RL^{2}\) algorithm and PPO in the `outer-loop'. We used three tasks from Tier-1
and trained each agent for 50 million environment steps using identical hyperparameters except for GRU size.  
Our meta-training approach was simplified compared to Meta-world, see Section~S.\ref{supp:simplifications}. 

\subsubsection{Bi-hemispheric training}
For the main experiment, we selected nine Meta-world tasks from Tier-1, 2 and 3, disjoint from Meta-training sub-tasks, to train and evaluate bi-hemispheric agents and the left-only baseline. These tasks exhibit varying degrees of novelty from the tasks used in meta-training. For each task, we trained for 5 million environment steps using the PPO algorithm \cite{schulman_proximal_2017}, as this timeframe allowed agents to learn each task. During training, we extracted mean rewards and median gating values from each batch. For bi-hemispheric agents, we also tested the left-hemisphere network as an independent agent. We made simplifications to training compared to Meta-world, following the Continual World approach \cite{wolczyk_continual_2021}; randomly sampling from 20 tasks (instead of 50) and making goal and object positions observable, see Section~S.\ref{supp:simplifications}.

\subsubsection{Hyperparameter selection} 
\label{sec:hyperparam}
For each task, we used identical hyperparameters for the left-only baseline and all bi-hemispheric agents. We selected hyperparameters that enabled best performance for the left-only baseline. This is a conservative approach that enabled us to evaluate bi-hemispheric agents against the left-only baselines at their best. Hyperparameter selection was informed by values used in Meta-world and previous studies into PPO settings \cite{yu_meta-world_2020,andrychowicz_what_2020,ni_recurrent_2022}. Gating network parameters were chosen for good final left-hemisphere performance. We also used Meta-world hyperparameter values to inform settings for meta-training. Hyperparameters are in given in Section~S.\ref{supp:hyperparams}.

\subsection{Main experiments} 
\label{sec:evaluation}

We compared bi-hemispheric agents against baselines using `relative rewards'; the ratio between rewards of two agents.
A relative reward of \(\geq 1\) indicates superior performance over the agent in the denominator.
To assess bi-hemispheric performance against Objectives 1 and 2, we placed bi-hemispheric agent reward in the numerator and the left-only baseline reward in the denominator. 

\subsubsection{Initial Relative Reward}
We created a metric called Initial~Relative~Reward~(IRR) to evaluate Objective 1. IRR measures whether a bi-hemispheric agent improves on initial performance of the left-only baseline. We calculated IRR using the ratio of median rewards of the first \(k = 1\) million environment steps for the bi-hemispheric agent over the left-only baseline:
\begin{equation}
    \label{eq:irr}
    IRR = \frac{Median(R_{t\leq k}^{bihem})}{Median(R_{t\leq k}^{left\;only})}
\end{equation}

\subsubsection{Final Relative Reward}
We created a metric called Final~Relative~Reward~(FRR) to evaluate Objective 2. Calculating FRR is similar to IRR, except we put rewards achieved by the left-hemisphere (on its own) in the numerator, and for the last \(k = 1\) million environment steps. \(T\) is the total number of environment steps.
\begin{equation}
    \label{eq:frr}
    FRR = \frac{Median(R_{t\geq T - k}^{left\;hemisphere})}{Median(R_{t\geq T - k}^{left\;only})}
\end{equation}

\section{Results} 
\label{sec:results}

This section focuses on the Main experiment, for meta-training, see Section~S.\ref{supp:metares}. 
For reference, per-step rewards for Meta-world tasks are between 0 and 10 where 10 occurs when the agent successfully achieves the specified goal of the task \cite{yu_meta-world_2020}. 
Given our objectives, we focus on rewards. However, Meta-world tasks are often evaluated on how often an agent achieves task success, which we include in a range of additional results plots in Section~S.\ref{supp:main_exps}.

\begin{figure}[th!]
    \centering
    \includegraphics[width=0.8\textwidth]{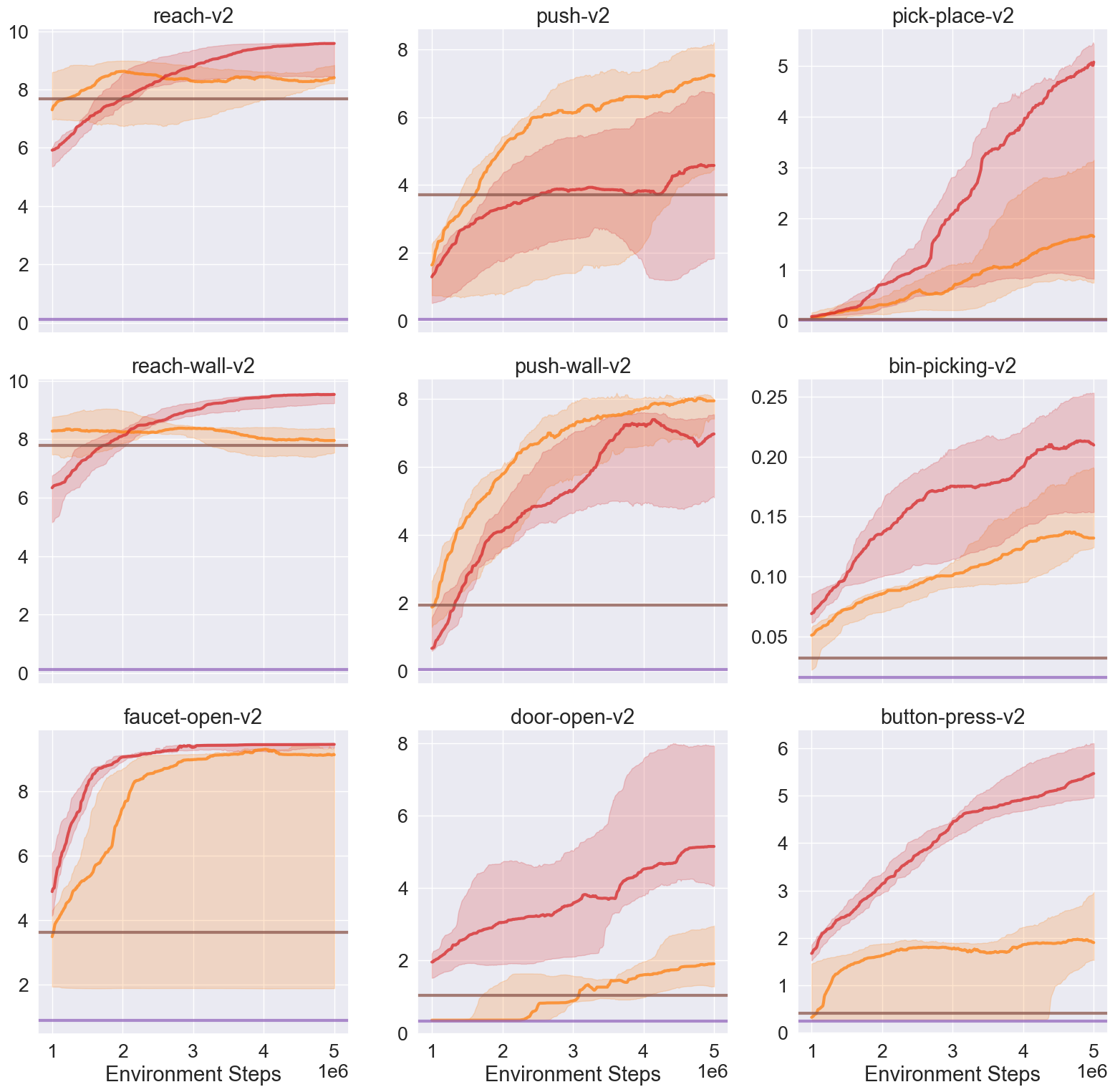}
    \includegraphics[scale=0.25]{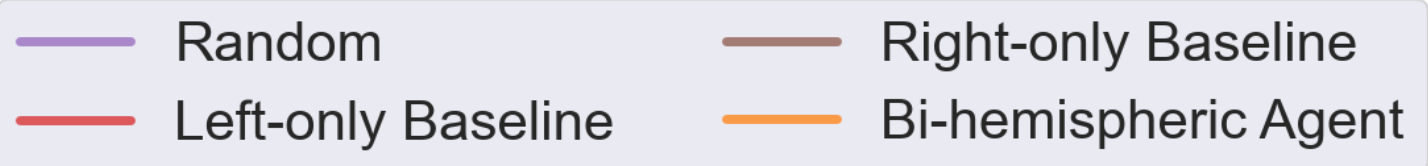}
    \caption{\small Training curves. The thick line indicates the median of all seed runs, while the shaded ribbon covers the range of min and max values.}
    \label{fig:bihem_mean_rew}
\end{figure}

Figure~\ref{fig:bihem_mean_rew} shows the mean rewards achieved for bi-hemispheric agents and baselines. 
For the bi-hemispheric agent and left-only baseline, mean rewards are smoothed using a rolling median over one million environment steps. Right-only and Random baseline median rewards are calculated by sampling sub-tasks with replacement. We used this approach as these baselines do not learn during evaluation, hence sampled tasks can be treated as independent.

We identify three groups of results. The first group consists of the tasks \texttt{reach-v2}, \texttt{push-v2}, \texttt{reach-wall-v2} and \texttt{push-wall-v2}. Here, the bi-hemispheric agent outperforms the left-only baseline initially and achieves comparable or better performance over the whole period. Tasks in this group are tasks where the right-hemisphere performs strongly.

The second group consists of \texttt{pick-place-v2} and \texttt{bin-picking-v2}. For these tasks, the left-only baseline generally outperforms the bi-hemispheric agent. In this group, overall rewards achieved for all agents are not very high, especially for \texttt{bin-picking-v2}, and vary across seeds. Notably, right-hemisphere performance is also poor on these tasks. This is consistent with Meta-world benchmark results, which show PPO agents fail to learn bin-picking, and that meta-learning agents take longer to learn \texttt{pick-place-v2} compared to \texttt{reach-v2} and \texttt{push-v2} \cite{yu_meta-world_2020}.

The third group contains the Tier-3 tasks. Here too, the left-only baseline outperforms the bi-hemispheric agent. Bi-hemispheric agent performance can vary significantly across seeds, and in some cases fail to learn the task at all. \texttt{Faucet-open-v2} is particularly affected by this, with many runs performing similarly to the left-only baseline while others failed to learn at all. Essentially, bi-hemispheric agents struggle with this degree of novelty.

\subsubsection{Objective 1: Initial performance} 
\label{sec:objective1}

\begin{figure}
    \centering
    \includegraphics[width=0.9\textwidth]{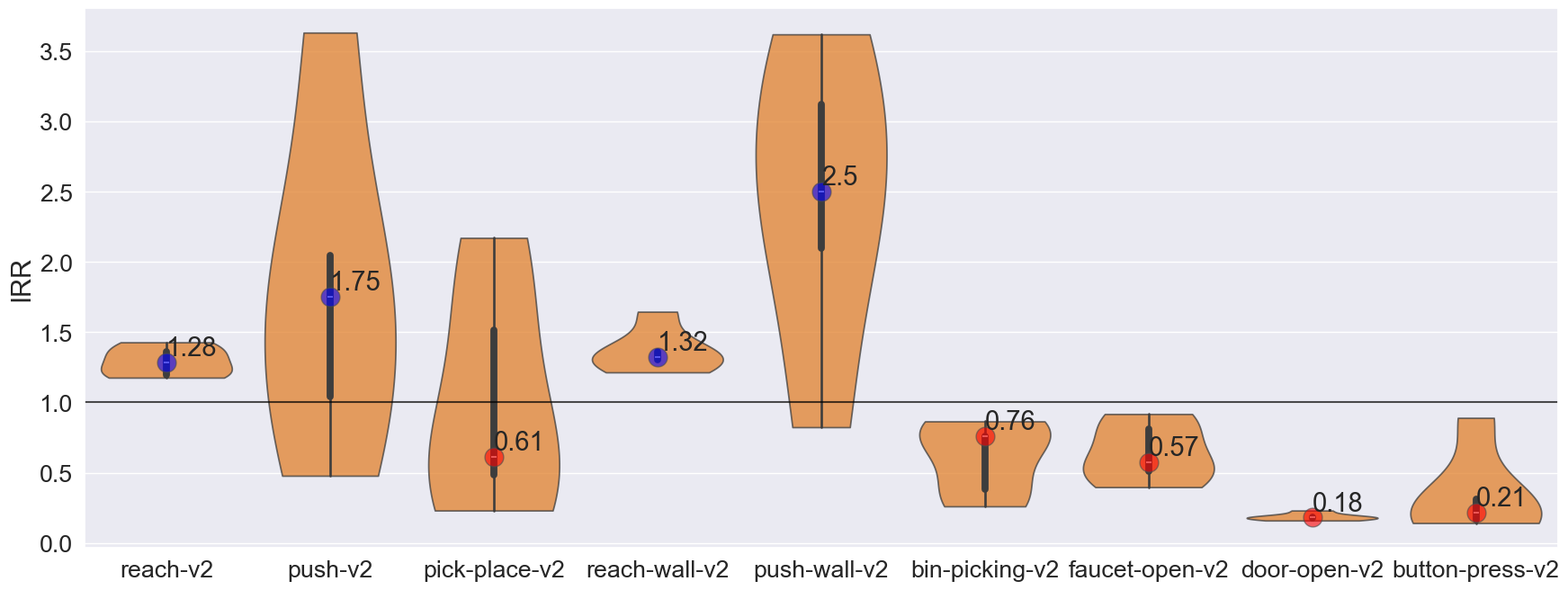}
    \caption{\small Initial bi-hemispheric agent performance relative to left-only baseline: IRR distribution over seeds. IRR scores of greater than 1 indicate that initial bi-hemispheric performance exceeds the left-only baseline. Median IRR scores which exceed 1 are shown in blue, while those below 1 are in red. The black line at \(IRR=1\) indicates parity with left-only baseline performance.}
    \label{fig:irr_results}
\end{figure}

Figure~\ref{fig:irr_results} plots IRR over tasks and seeds. 
Consistent with Figure~\ref{fig:bihem_mean_rew} we see that the bi-hemispheric agent achieves IRR scores larger than 1 for the reach and push families of tasks across different seeds. For other tasks, bi-hemispheric performance does not exceed the left-only baseline. For \texttt{pick-place-v2} on some seeds the bi-hemispheric agent outperforms the left-only baseline. However, given the initial rewards achieved for \texttt{pick-place-v2} are small, this may drive the differences in IRR scores, rather than genuinely improved performance.

When the right-hemisphere is competent, bi-hemispheric agents could avoid the poor initial outcomes associated with training an agent from scratch. This indicates that our agent design could exploit generalist skills in the right-hemisphere, when they exist, to avoid poor initial performance.

\subsubsection{Objective 2: Hemispheric shift} 
\label{sec:objective2}

\begin{figure}
    \centering
    \includegraphics[width=0.9\textwidth]{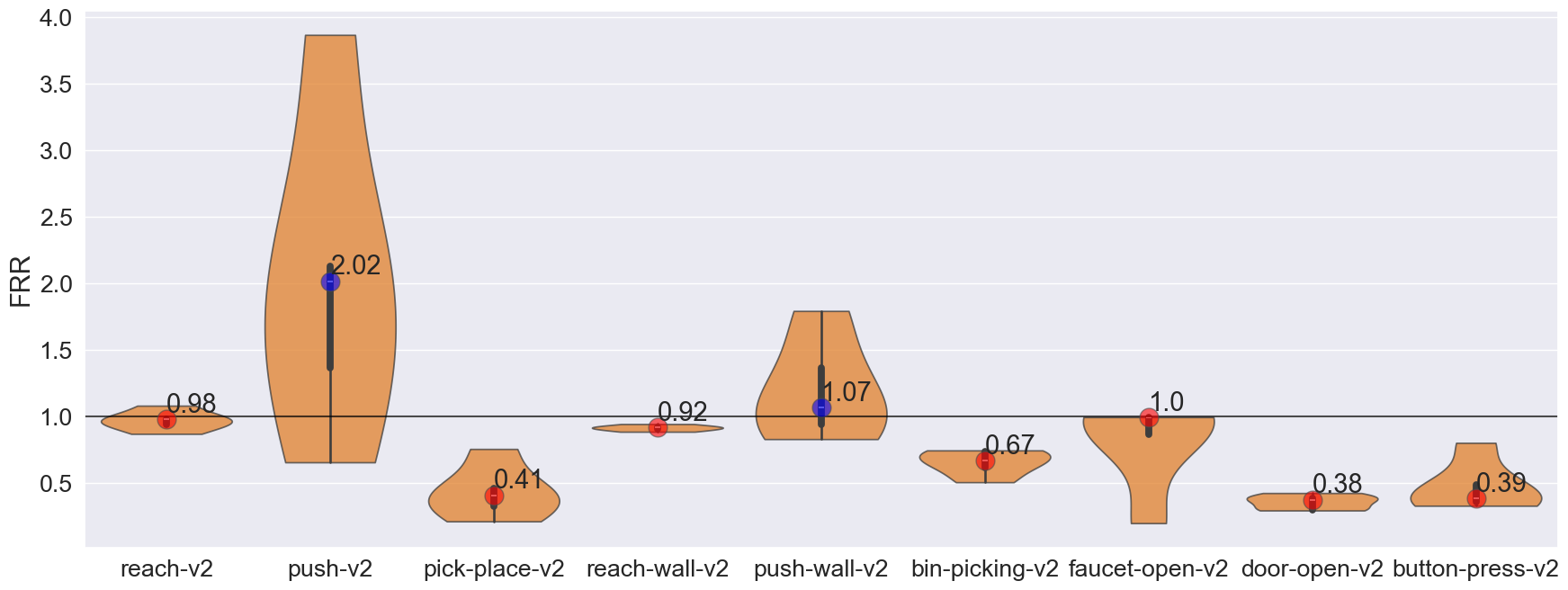}
    \caption{\small Final left-hemisphere performance relative to left-only baseline: FRR distribution over seeds.}
    \label{fig:frr_results}
\end{figure}

Figure~\ref{fig:frr_results} plots FRR across tasks and seeds. Scores greater than 1 indicate that the final performance of the left hemisphere exceeded the final performance of the left-only baseline. The bi-hemispheric agent only achieved FRR of greater than 1 on \texttt{push-v2} and \texttt{push-wall-v2}. The bi-hemispheric agent achieved FRR close to 1 for reach tasks and for \texttt{faucet-open-v2}. We observed FRR scores well below 1 for other tasks.

Overall, when the right-hemisphere performs well, we observed consistently higher FRR scores. This suggests that the right-hemisphere may have contributed to learning by teaching the left-hemisphere. Consequently, FRR scores may not need to exceed 1 for a bi-hemispheric approach to be worthwhile. If improvements to initial performance achieved by a bi-hemispheric agent are large enough, small declines in final left-hemisphere performance may be acceptable. 

\subsubsection{Combining Objectives 1 and 2}

\begin{figure}
    \centering
        \centering
        \includegraphics[scale=0.5]{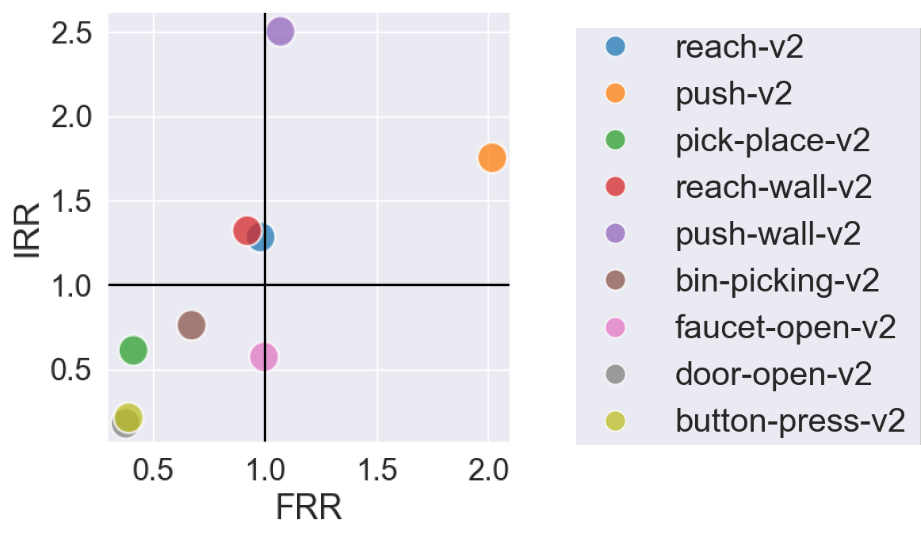}
    \caption{\small Comparison of IRR vs FRR. Black lines at \(IRR=1\) and \(FRR=1\) divide the plot into quadrants. The upper right quadrant is the ideal outcome where IRR and FRR exceed left-only baseline performance and the bi-hemispheric agent achieves Objectives 1 and 2. The lower-left quadrant indicates failure to achieve either objective.}
    \label{fig:irr_vs_frr}
\end{figure}

Figure~\ref{fig:irr_vs_frr} plots median IRR and FRR scores.
Only \texttt{push-v2} and \texttt{push-wall-v2} fell within the upper right quadrant. \texttt{Reach-v2} and \texttt{reach-wall-v2} achieve IRR scored greater than 1, but have FRR scores just below 1. However, all other tasks were located in the lower left quadrant. What differentiated the tasks in the bottom-left quadrant from those near to the upper-right is strong right-hemisphere performance. Essentially, when right-hemisphere performed tasks, bi-hemispheric agents achieved stronger IRR and FRR scores. 

\section{Discussion} 
\label{sec:discussion}

The key finding is that when the right-hemisphere contains generalist skills relevant to a task, the agent can exploit them to improve initial performance on that task. Furthermore, doing so does not significantly impact ability to learn the task. This was achieved for tasks which exhibit non-parametric variation from the meta-training tasks. However, this was not so for most tasks. Without a competent right-hemisphere, bi-hemispheric agents generally exhibit worse initial performance \emph{and} struggle to learn tasks to the same degree as an agent trained from scratch. This may be partially explained by poor right-hemisphere performance rather than the bi-hemispheric agent design. We discuss potential solutions below.

\subsubsection{Improved meta-training}
 Our approach to meta-training the right-hemisphere involved several simplifications (Section~S.\ref{supp:simplifications}), including training on fewer tasks and for fewer environment steps. Increasing either of these may result in more robust right-hemispheres with greater generalisation capacity.  Additionally, we saw doubling the GRU size in our right-only baseline resulted in large improvements in performance. Consequently, improving the right-hemisphere meta-training process may have significant benefits for bi-hemispheric agent performance.

\subsubsection{Training the left-hemisphere and gating network separately}
Particularly for the Tier-3 tasks, we observed training trajectories that were quite variable (see Figure~\ref{fig:bihem_mean_rew}). Some trajectories demonstrated reasonable growth while others exhibited long flat periods without any improvement. We hypothesise that this is due to interference between gating values and left-hemisphere gradients. Essentially, left-hemisphere gradients are scaled down by the gating values, which act as a dynamic learning rate. This may make learning a novel task difficult for the left-hemisphere if the right-hemisphere performs poorly.

On solution could be to separate the training of the gating network and left-hemisphere. This could be achieved by training each network with an off-policy algorithm. Off-policy algorithms use different policies for exploration and exploitation \cite{sutton_reinforcement_2018}. Consequently, the bi-hemispheric agent's joint policy would be used to explore while the left-hemisphere and gating network could each be trained separately, using importance sampling to treat the bi-hemispheric agent's decisions as their own. 

\subsubsection{Extension to continual learning}
Applying bi-hemispheric agents in continual learning may be where their capabilities become more beneficial. If agents avoid poor initial performance on novel tasks, yet still learn, it would result in better overall performance. For our agent to operate in this setting, it should store learnt policies in the left-hemisphere, infer whether a task has an existing policy or is novel, and retrieve existing policies for previously encountered tasks from the left-hemisphere. We leave this to future work, but note that the adaptive MoE proposed by Tsuda et al. \cite{tsuda_modeling_2020} may help address the issue of task retrieval.
They used gating networks to select which expert to use for novel tasks. This design could be applied as a left-hemisphere alongside a meta-trained right-hemisphere.


\section{Conclusion} 

We developed a novel RL-agent design with specialised brain hemispheres as per the NRH. Our agent can exploit generalist skills, when present, from the right-hemisphere to improve initial performance over an agent trained from scratch, with minimal impact on its ability to learn novel tasks. Despite this, when the right-hemisphere lacks these skills, our agent struggles to perform novel tasks. We also identified potential improvements and extensions to enable the agent to operate in continual settings.

\bibliographystyle{splncs04}
\bibliography{references}

\begin{thebibliography}{10}
\providecommand{\url}[1]{\texttt{#1}}
\providecommand{\urlprefix}{URL }
\providecommand{\doi}[1]{https://doi.org/#1}

\bibitem{andrychowicz_what_2020}
Andrychowicz, M., Raichuk, A., Stańczyk, P., Orsini, M., Girgin, S., Marinier, R., Hussenot, L., Geist, M., Pietquin, O., Michalski, M., Gelly, S., Bachem, O.: What {Matters} for {On}-{Policy} {Deep} {Actor}-{Critic} {Methods}? {A} {Large}-{Scale} {Study} (Oct 2020)

\bibitem{botvinick_reinforcement_2019}
Botvinick, M., Ritter, S., Wang, J.X., Kurth-Nelson, Z., Blundell, C., Hassabis, D.: Reinforcement {Learning}, {Fast} and {Slow}. Trends in Cognitive Sciences  \textbf{23}(5),  408--422 (May 2019)

\bibitem{chang_unified_2020}
Chang, Y.N., Lambon~Ralph, M.A.: A unified neurocomputational bilateral model of spoken language production in healthy participants and recovery in poststroke aphasia. Proceedings of the National Academy of Sciences of the United States of America  \textbf{117}(51),  32779--32790 (Dec 2020)

\bibitem{duan_rl2_2016}
Duan, Y., Schulman, J., Chen, X., Bartlett, P.L., Sutskever, I., Abbeel, P.: {RL}\${\textasciicircum}2\$: {Fast} {Reinforcement} {Learning} via {Slow} {Reinforcement} {Learning} (2016)

\bibitem{finn_model-agnostic_2017}
Finn, C., Abbeel, P., Levine, S.: Model-{Agnostic} {Meta}-{Learning} for {Fast} {Adaptation} of {Deep} {Networks}. In: Precup, D., Teh, Y.W. (eds.) Proceedings of the 34th {International} {Conference} on {Machine} {Learning}. Proceedings of {Machine} {Learning} {Research}, vol.~70, pp. 1126--1135. PMLR (Aug 2017)

\bibitem{gazzaniga_cerebral_2000}
Gazzaniga, M.S.: Cerebral specialization and interhemispheric communication: {Does} the corpus callosum enable the human condition? Brain  \textbf{123}(7),  1293--1326 (Jul 2000)

\bibitem{goldberg_hemisphere_1981}
Goldberg, E., Costa, L.D.: Hemisphere differences in the acquisition and use of descriptive systems. Brain and Language  \textbf{14}(1),  144--173 (Sep 1981)

\bibitem{goldberg_lateralization_1994}
Goldberg, E., Podell, K., Lovell, M.: Lateralization of frontal lobe functions and cognitive novelty. The Journal of Neuropsychiatry and Clinical Neurosciences  \textbf{6}(4),  371--378 (Nov 1994)

\bibitem{goldberg_hemispheric_2013}
Goldberg, E., Roediger, D., Kucukboyaci, N.E., Carlson, C., Devinsky, O., Kuzniecky, R., Halgren, E., Thesen, T.: Hemispheric asymmetries of cortical volume in the human brain. Cortex  \textbf{49}(1),  200--210 (Jan 2013)

\bibitem{hassabis_neuroscience-inspired_2017}
Hassabis, D., Kumaran, D., Summerfield, C., Botvinick, M.: Neuroscience-{Inspired} {Artificial} {Intelligence}. Neuron  \textbf{95}(2),  245--258 (Jul 2017)

\bibitem{jacobs_adaptive_1991}
Jacobs, R.A., Jordan, M.I., Nowlan, S.J., Hinton, G.E.: Adaptive {Mixtures} of {Local} {Experts}. Neural Computation  \textbf{3}(1),  79--87 (Feb 1991)

\bibitem{khetarpal_towards_2022}
Khetarpal, K., Riemer, M., Rish, I., Precup, D.: Towards {Continual} {Reinforcement} {Learning}: {A} {Review} and {Perspectives}. Journal of Artificial Intelligence Research  \textbf{75},  1401--1476 (Dec 2022)

\bibitem{kirk_survey_2023}
Kirk, R., Zhang, A., Grefenstette, E., Rocktäschel, T.: A {Survey} of {Zero}-shot {Generalisation} in {Deep} {Reinforcement} {Learning}. Journal of Artificial Intelligence Research  \textbf{76},  201--264 (Jan 2023)

\bibitem{mnih_asynchronous_2016}
Mnih, V., Badia, A.P., Mirza, M., Graves, A., Lillicrap, T., Harley, T., Silver, D., Kavukcuoglu, K.: Asynchronous {Methods} for {Deep} {Reinforcement} {Learning}. In: Proceedings of {The} 33rd {International} {Conference} on {Machine} {Learning}. pp. 1928--1937. PMLR (Jun 2016), iSSN: 1938-7228

\bibitem{monaghan_hemispheric_2004}
Monaghan, P.: Hemispheric asymmetries in the split-fovea model of semantic processing. Brain and Language  \textbf{88}(3),  339--354 (Mar 2004)

\bibitem{ni_recurrent_2022}
Ni, T., Eysenbach, B., Salakhutdinov, R.: Recurrent {Model}-{Free} {RL} {Can} {Be} a {Strong} {Baseline} for {Many} {POMDPs}. In: Proceedings of the 39th {International} {Conference} on {Machine} {Learning}. pp. 16691--16723. PMLR (Jun 2022), iSSN: 2640-3498

\bibitem{ortega_meta-learning_2019}
Ortega, P.A., Wang, J.X., Rowland, M., Genewein, T., Kurth-Nelson, Z., Pascanu, R., Heess, N., Veness, J., Pritzel, A., Sprechmann, P., Jayakumar, S.M., McGrath, T., Miller, K., Azar, M., Osband, I., Rabinowitz, N., György, A., Chiappa, S., Osindero, S., Teh, Y.W., van Hasselt, H., de~Freitas, N., Botvinick, M., Legg, S.: Meta-learning of {Sequential} {Strategies} (2019)

\bibitem{parisi_continual_2019}
Parisi, G.I., Kemker, R., Part, J.L., Kanan, C., Wermter, S.: Continual lifelong learning with neural networks: {A} review. Neural Networks  \textbf{113},  54--71 (May 2019)

\bibitem{peleg_two_2010}
Peleg, O., Manevitz, L., Hazan, H., Eviatar, Z.: Two hemispheres—two networks: a computational model explaining hemispheric asymmetries while reading ambiguous words. Annals of Mathematics and Artificial Intelligence  \textbf{59}(1),  125--147 (May 2010)

\bibitem{pham_dualnet_2021}
Pham, Q., Liu, C., Hoi, S.: {DualNet}: {Continual} {Learning}, {Fast} and {Slow} (Nov 2021)

\bibitem{rajagopalan_deep_2022}
Rajagopalan, C., Rawlinson, D., Goldberg, E., Kowadlo, G.: Deep learning in a bilateral brain with hemispheric specialization (2022)

\bibitem{schulman_proximal_2017}
Schulman, J., Wolski, F., Dhariwal, P., Radford, A., Klimov, O.: Proximal {Policy} {Optimization} {Algorithms} (2017)

\bibitem{shevtsova_neural_1999}
Shevtsova, N., Reggia, J.A.: A {Neural} {Network} {Model} of {Lateralization} during {Letter} {Identification}. Journal of Cognitive Neuroscience  \textbf{11}(2),  167--181 (Mar 1999)

\bibitem{sutton_reinforcement_2018}
Sutton, R.S., Barto, A.G.: Reinforcement learning: {An} introduction. MIT Press (2018)

\bibitem{tsuda_modeling_2020}
Tsuda, B., Tye, K.M., Siegelmann, H.T., Sejnowski, T.J.: A modeling framework for adaptive lifelong learning with transfer and savings through gating in the prefrontal cortex. Proceedings of the National Academy of Sciences  \textbf{117}(47),  29872--29882 (Nov 2020)

\bibitem{wang_learning_2016}
Wang, J.X., Kurth-Nelson, Z., Tirumala, D., Soyer, H., Leibo, J.Z., Munos, R., Blundell, C., Kumaran, D., Botvinick, M.: Learning to reinforcement learn (2016)

\bibitem{weems_hemispheric_2004}
Weems, S.A., Reggia, J.A.: Hemispheric specialization and independence for word recognition: {A} comparison of three computational models. Brain and Language  \textbf{89}(3),  554--568 (Jun 2004)

\bibitem{wolczyk_continual_2021}
Wolczyk, M., Zając, M., Pascanu, R., Kuciński, L., Miloś, P.: Continual {World}: {A} {Robotic} {Benchmark} {For} {Continual} {Reinforcement} {Learning}. In: Advances in {Neural} {Information} {Processing} {Systems}. vol.~34, pp. 28496--28510. Curran Associates, Inc. (2021)

\bibitem{yu_meta-world_2020}
Yu, T., Quillen, D., He, Z., Julian, R., Hausman, K., Finn, C., Levine, S.: Meta-{World}: {A} {Benchmark} and {Evaluation} for {Multi}-{Task} and {Meta} {Reinforcement} {Learning}. In: Proceedings of the {Conference} on {Robot} {Learning}. pp. 1094--1100. PMLR (May 2020), iSSN: 2640-3498

\bibitem{zintgraf_varibad_2021}
Zintgraf, L., Schulze, S., Lu, C., Feng, L., Igl, M., Shiarlis, K., Gal, Y., Hofmann, K., Whiteson, S.: {VariBAD}: {Variational} {Bayes}-{Adaptive} {Deep} {RL} via {Meta}-{Learning}. Journal of Machine Learning Research  \textbf{22}(289),  1--39 (2021)

\end{thebibliography}


\begin{thebibliography}{1}
\providecommand{\url}[1]{\texttt{#1}}
\providecommand{\urlprefix}{URL }
\providecommand{\doi}[1]{https://doi.org/#1}

\bibitem{schulman_proximal_2017}
Schulman, J., Wolski, F., Dhariwal, P., Radford, A., Klimov, O.: Proximal {Policy} {Optimization} {Algorithms} (2017)

\bibitem{wolczyk_continual_2021}
Wolczyk, M., Zając, M., Pascanu, R., Kuciński, L., Miloś, P.: Continual {World}: {A} {Robotic} {Benchmark} {For} {Continual} {Reinforcement} {Learning}. In: Advances in {Neural} {Information} {Processing} {Systems}. vol.~34, pp. 28496--28510. Curran Associates, Inc. (2021)

\bibitem{yu_meta-world_2020}
Yu, T., Quillen, D., He, Z., Julian, R., Hausman, K., Finn, C., Levine, S.: Meta-{World}: {A} {Benchmark} and {Evaluation} for {Multi}-{Task} and {Meta} {Reinforcement} {Learning}. In: Proceedings of the {Conference} on {Robot} {Learning}. pp. 1094--1100. PMLR (May 2020), iSSN: 2640-3498

\end{thebibliography}

\end{document}


\title{Supplementary information for ``Graceful task adaptation with a bi-hemispheric RL agent''}

\titlerunning{Supplementary information}

\author{}
\authorrunning{}
\institute{}

\maketitle

\section{Agent design}
\label{supp:agents}

\subsection{Network architecture}

\begin{figure}
   \centering
   \includegraphics[width=0.85\textwidth]{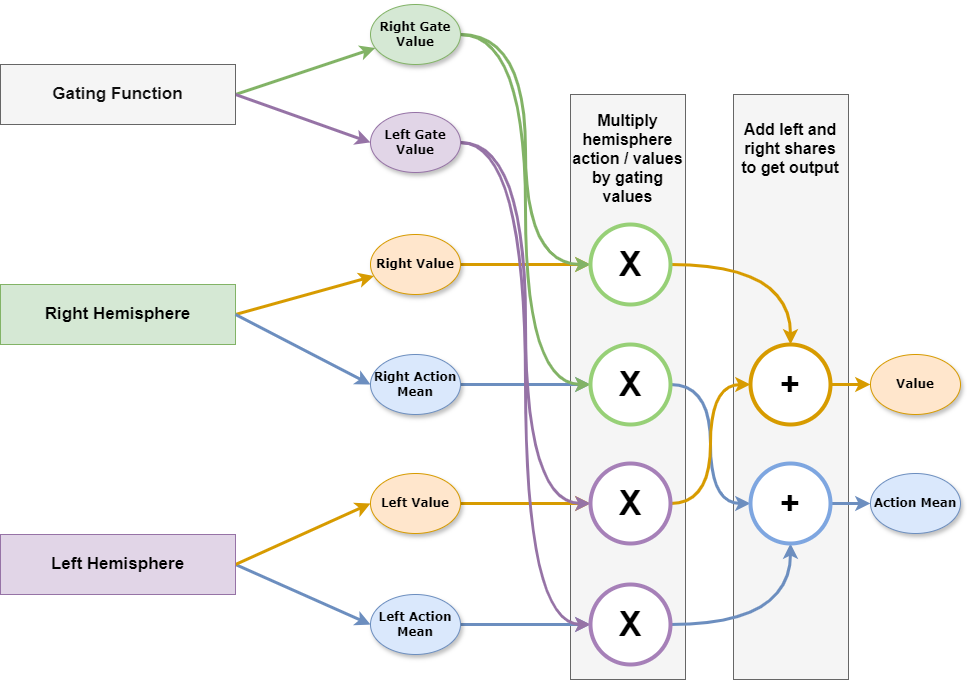}
   \caption{Bi-hemispheric agent architecture. There is a network for left hemisphere, right hemisphere and the gating network. The gating function combines action and value estimates with `gating values' to assign `responsibility' to a hemisphere.}
   \label{fig:network_arch}
 \end{figure}

A diagram of the network architecture is shown in Figure~\ref{fig:network_arch}.

\subsection{Loss function}
\label{supp:loss}
\begin{figure}[h]
  \centering
  \includegraphics[width=0.5\textwidth]{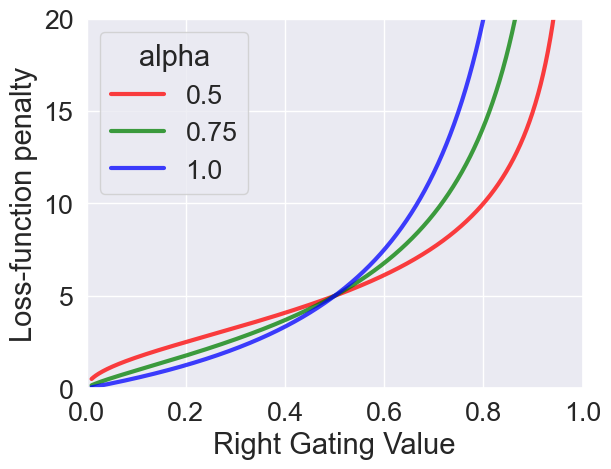}
  \caption{Example penalty curves (\(\beta = 5\))}
  \label{fig:penalty_curves}
\end{figure}

We add an additive penalty to the bi-hemispheric agent's loss function to encourage hemispheric shift. In this case, we use the standard PPO loss \cite{schulman_proximal_2017}:

\begin{equation}
    L^{bihem} = L^{PPO} +\beta(\frac{P^{right}}{P^{left}})^{\alpha}
\end{equation}
The penalty is large when \(P^{right}\) is significantly larger than \(P^{left}\) and approaches zero as \(P^{left}\) grows. \(\beta\) controls the strength of the penalty, while \(\alpha\) influences the slope of the penalty with respect to \(P^{right}\). We show examples in Figure~\ref{fig:penalty_curves}.

\section{Experiments}

\subsection{Tasks}
\label{supp:tasks}

The experiment environments are shown in Table~\ref{tab:environment_table}. While there does exist a \texttt{pick-place-wall-v2} task in Meta-world, we found that regardless of action, rewards would always be zero, leading us to suspect an error. Consequently, we chose \texttt{bin-picking-v2} for our Tier-2 equivalent to \texttt{pick-place-v2}.

 \begin{table}[h!]
  \centering
    \begin{tabular}{ccc}
    \toprule 
    \textbf{Tier} & \textbf{Environment name} & \textbf{Objective} \\
    \midrule 
    {Tier-1} & \texttt{reach-v2} & Reach location\\
    & \texttt{push-v2} & Push puck to location\\
    & \texttt{pick-place-v2} & Pick up puck and place in location\\
    \bottomrule \\
    \multirow{3}{4em}{Tier-2} & reach-wall-v2 & Reach location and bypass wall\\
    & \texttt{push-wall-v2} & Push puck to location and bypass wall\\
    & \texttt{bin-picking-v2} & Pick up puck from bin and place in another bin\\
    \bottomrule \\
    \multirow{3}{4em}{Tier-3} & \texttt{faucet-open-v2} & Rotate faucet counter-clockwise\\
    & \texttt{door-open-v2} & Open door with revolving joint\\
    & \texttt{button-press-v2} & Press button\\
    \bottomrule
    \end{tabular}
    \caption{Experiment environments}
    \label{tab:environment_table}
\end{table}

\subsection{Baselines}
A list of all agents and baselines is in Table~\ref{tab:experiment_table} and GRU sizes are in Table~\ref{tab:network_size}. 

\begin{table}
  \centering
    \setlength{\tabcolsep}{6pt} 
    \begin{tabular}{m{5em} m{15em} m{15em}}
    \toprule \textbf{Algorithm} & \textbf{Description} & \textbf{Objective} \\
    \midrule Random & Randomly selects actions & Determine lower bound of performance \\
    \midrule
    Right-only & Meta-trained agent with double weights of right-hemisphere & Evaluate meta-training only \\
    \hline
    Left-only & Agent trained from scratch with double weights of left-hemisphere & Evaluate training from scratch \\
	\hline
    Bi-hemispheric & Agent with meta-trained right-hemisphere and randomly initialised left-hemisphere and gating network & Evaluate impact of bi-hemispheric design \\
    \bottomrule
    \end{tabular}
    \caption{Agents}
    \label{tab:experiment_table}
\end{table}

\begin{table}
  \centering
    \begin{tabular}{m{10em} m{5em} m{5em}}
    \toprule 
    \textbf{Network} & \textbf{GRU size} & \textbf{Policy head size} \\
    \midrule
    Right-only & 256 & 512\\
    Left-only & 256 & 512\\
    \midrule
    Left hemisphere & 128 & 512\\
    Right hemisphere & 128 & 512\\
    \bottomrule
    \end{tabular}
    \caption{Network sizes}
    \label{tab:network_size}
\end{table}

\subsection{Experiment simplifications}
\label{supp:simplifications}

We simplified the meta-training process from the approach used in Meta-world benchmarks by shortening the training time and the number of training tasks. Published Meta-world results are trained for over 300 million environment steps \cite{yu_meta-world_2020}. This was impractical within our timeframes, although it may improve meta-trained agent performance. Additionally, Meta-world benchmarks use 1, 10 or 45 tasks \cite{yu_meta-world_2020}. We chose 3 tasks to enable generalisation across tasks, but avoid difficulties with training on a larger group of tasks.

The main experiment was also simplified compared to Meta-world, to reduce training time, based on the approach used in Continual World \cite{wolczyk_continual_2021} and using a wrapper provided by the authors. Firstly, we reduced the number of sub-tasks used during the bi-hemispheric training process; randomly sampling one of 20 available sub-tasks upon reset (instead of 50 in Meta-world). We also made goal and object positions observable, which made training easier and meant that each task was slightly different (i.e. more novel) from meta-training tasks, as the agent could draw on information about the objective.

\subsection{Hyperparameters} 
\label{supp:hyperparams}

Table~\ref{tab:lo_bihem_hp} shows hyperparameters used to train bi-hemispheric agents and left-only baseline agents.
Table~\ref{tab:gating_hp} shows the selected values of \(\alpha\) and \(\beta\) for the additive loss term for the bi-hemispheric agent.
Table~\ref{tab:metatraining_hp} shows hyper-parameters used for meta-training the right-hemisphere and right-only baseline.

\begin{table}
\begin{adjustwidth}{-1.in}{0in}
\tiny
\centering
\begin{tabular}{|l|c|c|c|c|c|c|c|c|c|}
\hline
 & \textbf{reach-v2} & \textbf{push-v2} & \textbf{pick-place-v2} & \textbf{reach-wall-v2} & \textbf{push-wall-v2} & \textbf{bin-picking-v2} & \textbf{faucet-open-v2} & \textbf{door-open-v2} & \textbf{button-press-v2} \\ \hline
\textbf{Learning rate} & 1.00E-05 & 1.00E-04 & 1.00E-04 & 1.00E-05 & 1.00E-04 & 1.00E-04 & 1.00E-05 & 1.00E-05 & 1.00E-05 \\ \hline
\textbf{Entropy coefficient} & 1.00E-05 & 1.00E-05 & 1.00E-05 & 1.00E-05 & 1.00E-05 & 1.00E-05 & 1.00E-05 & 1.00E-05 & 1.00E-05 \\ \hline
\textbf{Discount rate} & 0.99 & 0.99 & 0.99 & 0.99 & 0.99 & 0.99 & 0.99 & 0.99 & 0.99 \\ \hline
\textbf{GAE lambda} & 0.97 & 0.97 & 0.97 & 0.97 & 0.97 & 0.97 & 0.97 & 0.9 & 0.97 \\ \hline
\textbf{Optimiser} & Adam & Adam & Adam & Adam & Adam & Adam & Adam & Adam & Adam \\ \hline
\textbf{Normalise rewards} & TRUE & TRUE & TRUE & TRUE & TRUE & TRUE & TRUE & TRUE & TRUE \\ \hline
\textbf{PPO clip param} & 0.2 & 0.2 & 0.2 & 0.2 & 0.2 & 0.2 & 0.2 & 0.2 & 0.2 \\ \hline
\textbf{PPO epochs} & 8 & 8 & 8 & 8 & 8 & 8 & 8 & 8 & 8 \\ \hline
\textbf{Batch size} & 20 & 20 & 20 & 20 & 20 & 20 & 20 & 20 & 20 \\ \hline
\end{tabular}
\caption{Left-only baseline and bi-hemispheric hyperparameters}
\label{tab:lo_bihem_hp}
\end{adjustwidth}
\end{table}

\begin{table}
\begin{adjustwidth}{-1.in}{0in}
\tiny
\centering
\begin{tabular}{|l|c|c|c|c|c|c|c|c|c|}
\hline
 & \textbf{reach-v2} & \textbf{push-v2} & \textbf{pick-place-v2} & \textbf{reach-wall-v2} & \textbf{push-wall-v2} & \textbf{bin-picking-v2} & \textbf{faucet-open-v2} & \textbf{door-open-v2} & \textbf{button-press-v2} \\ \hline
\textbf{Alpha} & 0.75 & 0.75 & 0.75 & 0.75 & 0.75 & 0.75 & 0.75 & 1 & 0.75 \\ \hline
\textbf{Beta} & 5 & 5 & 5 & 5 & 5 & 5 & 5 & 5 & 5 \\ \hline
\end{tabular}
\caption{Gating hyperparameters}
\label{tab:gating_hp}
\end{adjustwidth}
\end{table}

\begin{table}
\centering
\begin{tabular}{|c|c|}
\hline
 \textbf{Right-hemisphere} & \textbf{Right-only baseline} \\ \hline
\textbf{Episodes per trial} 10 & 10 \\ \hline
\textbf{Learning rate} 5.00E-04 & 5.00E-04 \\ \hline
\textbf{Entropy coefficient} 5.00E-06 & 5.00E-06 \\ \hline
\textbf{Discount rate} 0.99 & 0.99 \\ \hline
\textbf{Optimiser} Adam & Adam \\ \hline
\textbf{Normalise rewards} TRUE & TRUE \\ \hline
\textbf{PPO clip param} 0.2 & 0.2 \\ \hline
\textbf{PPO epochs} 10 & 10 \\ \hline
\end{tabular}
\caption{Meta-training hyperparameters}
\label{tab:metatraining_hp}
\end{table}

\section{Results} 

\subsection{Meta-training}
\label{supp:metares}

\begin{figure}
  \centering
  \begin{subfigure}[]{\textwidth}
    \centering
    \includegraphics[width=0.95\textwidth]{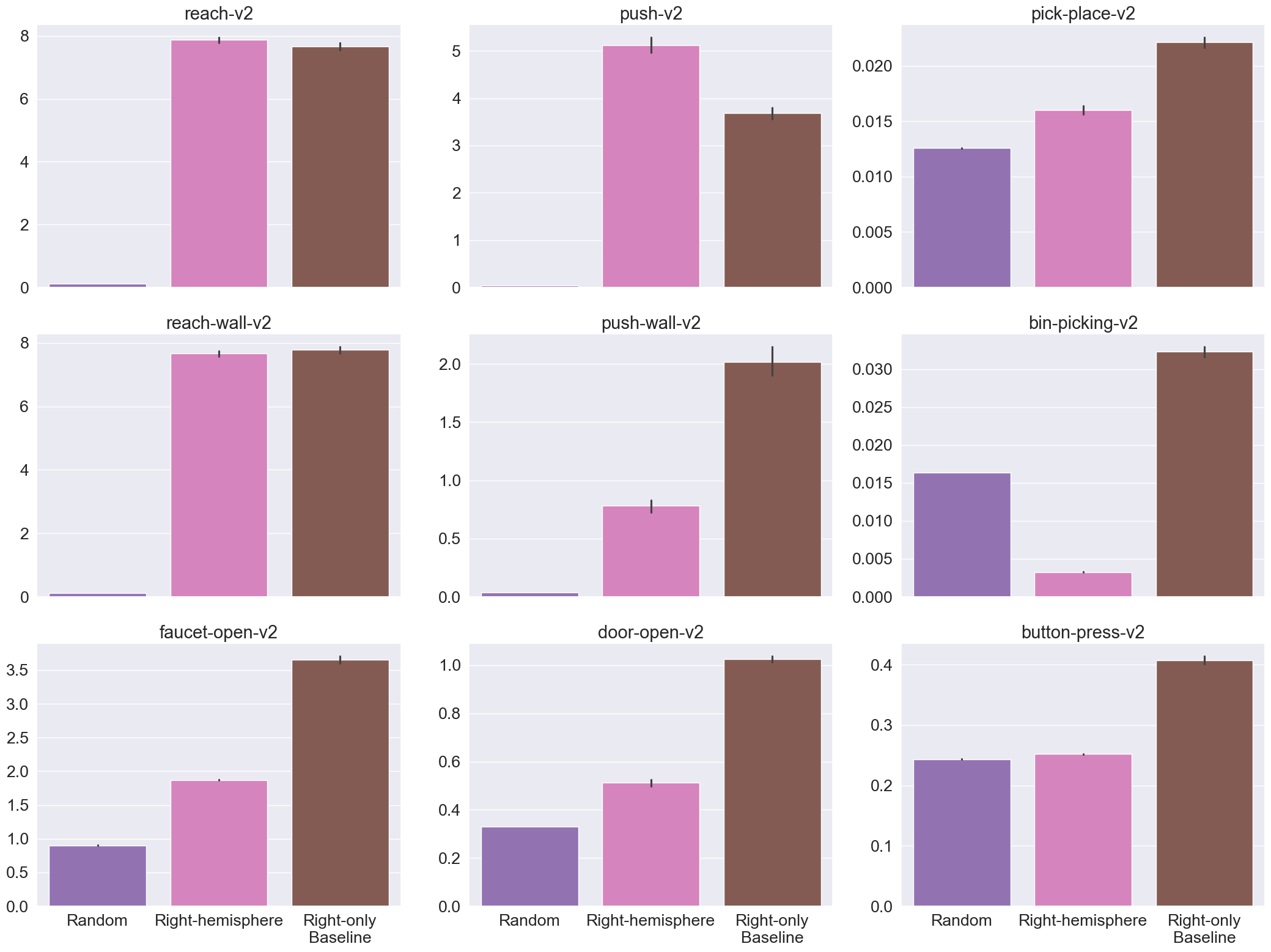}
    \caption{Average rewards}
  \end{subfigure}
  \begin{subfigure}[]{\textwidth}
    \centering
    \includegraphics[width=0.95\textwidth]{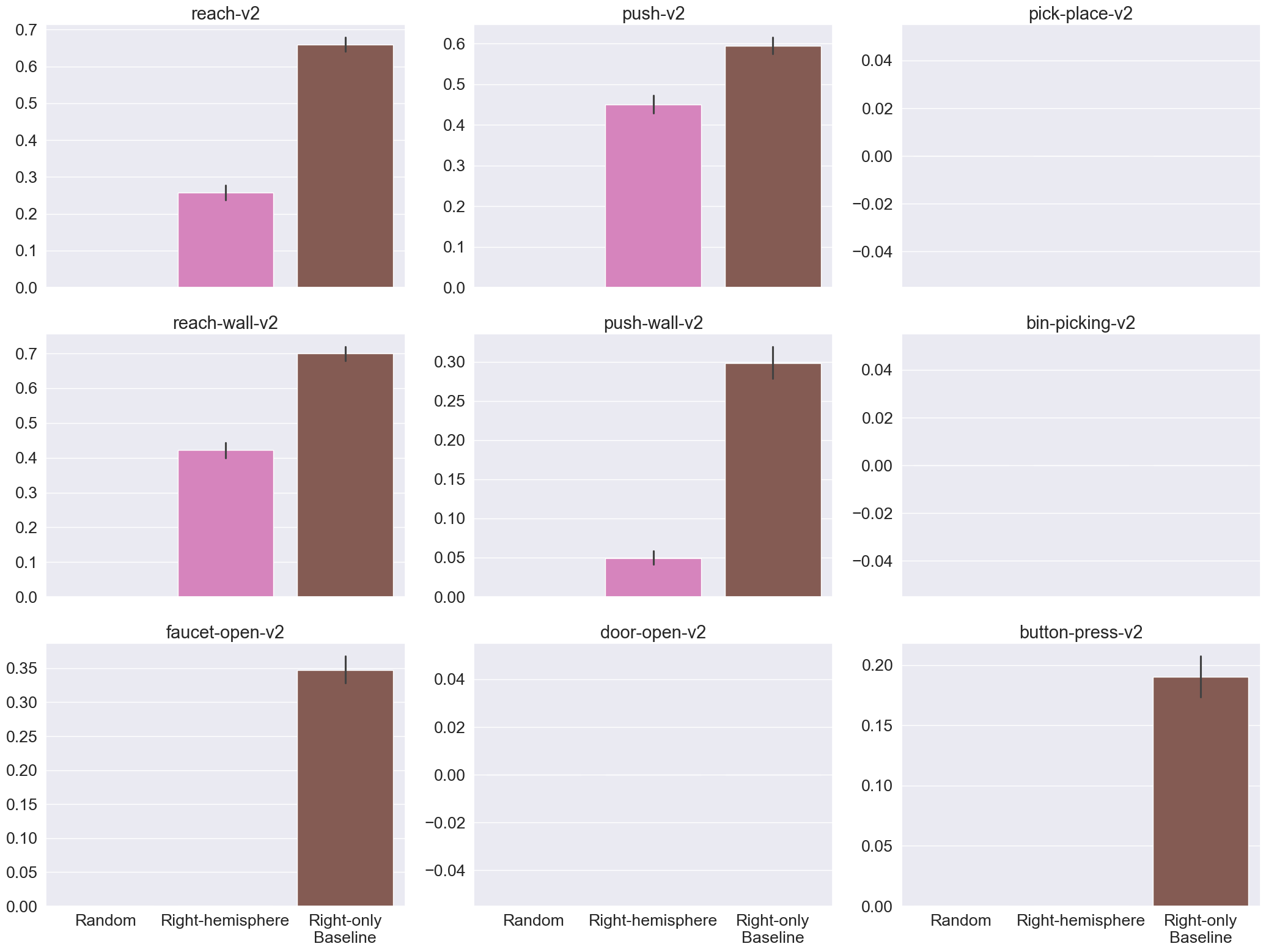}
    \caption{Success rate}
  \end{subfigure}
  \caption{\(RL^{2}\) meta-test results}
  \label{fig:rl2_metatest}
\end{figure}

\begin{figure}
  \centering
  \begin{subfigure}[]{\textwidth}
    \centering
    \includegraphics[width=0.95\textwidth]{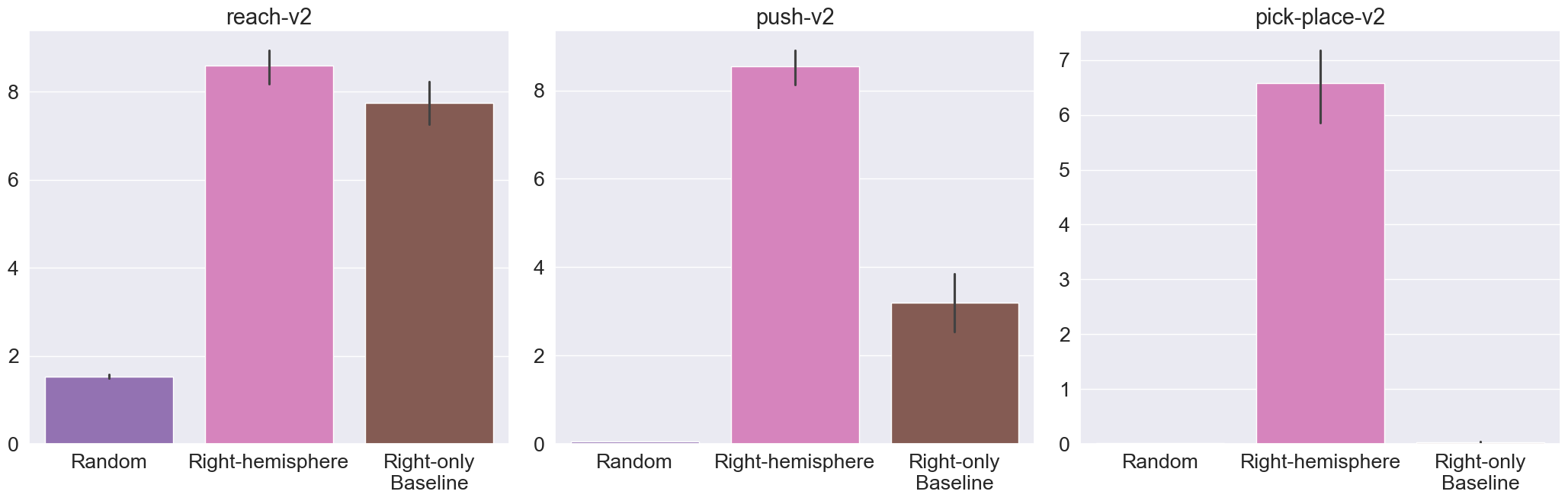}
    \caption{Average rewards}
  \end{subfigure}
  \begin{subfigure}[]{\textwidth}
      \centering
      \includegraphics[width=0.95\textwidth]{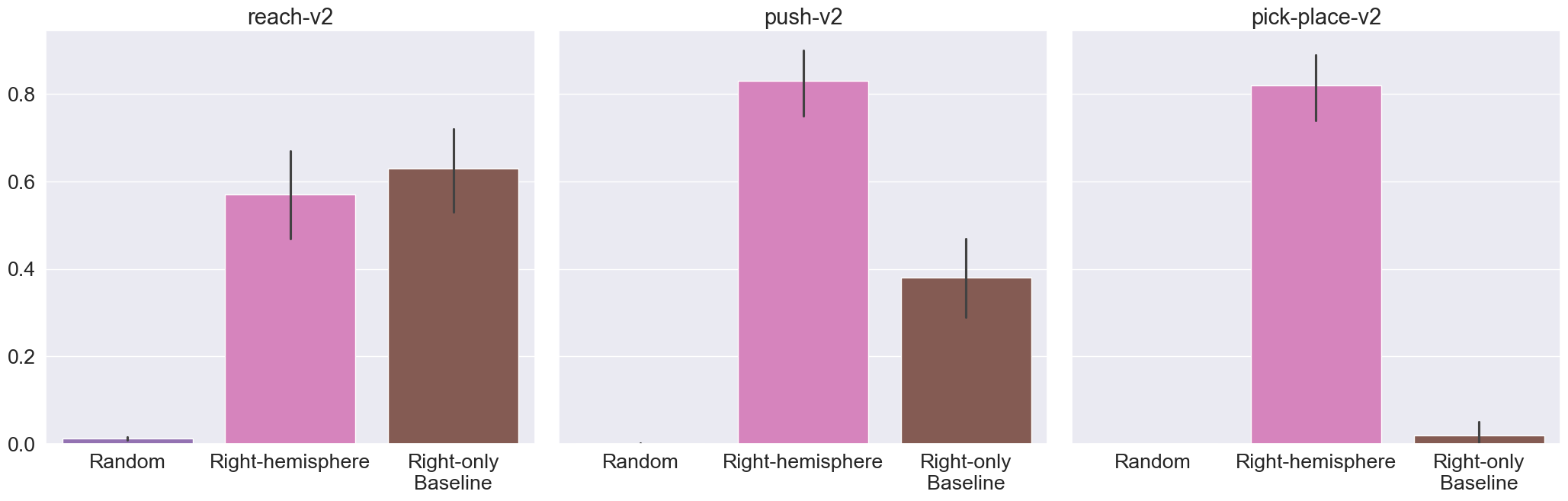}
      \caption{Success rate}
  \end{subfigure}
  \caption{\(RL^{2}\) meta-training results}
  \label{fig:rl2_metatrain}
\end{figure}

Figure~\ref{fig:rl2_metatest} shows the effectiveness of meta-training by showing the performance of the right-hemisphere and right-only baseline on the Main experiment tasks. 
Performance on the training set, the meta-training tasks, is shown in Figure~\ref{fig:rl2_metatrain}.
We generated these results by sampling sub-tasks from the evaluation tasks without replacement. Each agent was evaluated on 2,000 sampled sub-tasks to generate an estimate of average reward and success rate.

The \(RL^{2}\) agents could generalise to unseen tasks to some extent. For all tasks except \texttt{bin-picking-v2}, both \(RL^{2}\) agents exceeded a random agent. Performance was particularly strong for the \texttt{reach-v2} and \texttt{push-v2} tasks, and their Tier-2 counterparts; \texttt{reach-wall-v2} and \texttt{push-wall-v2}.

Both \(RL^{2}\) agents performed poorly on \texttt{pick-place-v2} and \texttt{bin-picking-v2}. This is despite \texttt{pick-place-v2}'s inclusion in the meta-training set. \texttt{Pick-place-v2} appears to be more difficult than \texttt{reach-v2} and \texttt{push-v2}. Left-only baselines achieved lower rewards on \texttt{pick-place-v2} than these tasks. Furthermore, learning curves in \citet{yu_meta-world_2020} indicate that \texttt{pick-place-v2} takes longer to learn than \texttt{reach-v2} and, to a lesser extent, \texttt{push-v2} (\cite{yu_meta-world_2020}). As the meta-training set combined these tasks, we anticipate that this might have resulted in agents learning the easier tasks faster and at the expense of \texttt{pick-place-v2}. This may also explain poor rewards on \texttt{bin-picking-v2}.

For Tier-3 tasks, we see that performance is better than random for both agents, but not to the same degree as \texttt{reach-v2} and \texttt{push-v2}. \texttt{Faucet-open-v2} achieved rewards of a similar order of magnitude as \texttt{push-v2}, while \texttt{door-open-v2} and \texttt{button-press-v2} achieved smaller rewards of less than one. Notably, the right-only baseline could achieve success on around 20\% and 30\% of the sampled tasks for \texttt{faucet-open-v2} and \texttt{button-press-v2} respectively, despite not having seen these tasks during training. This may indicate that greater generalisation could be achieved by increasing the number of weights in \(RL^{2}\) agents.

Overall, we found that both right-hemisphere and right-only baseline agents could generalise to novel tasks to some degree. Consequently, we could use the right-hemisphere agent to instil general skills into our bi-hemispheric agent. We note, however, that despite being Tier-1 and Tier-2 tasks -- our agents had substantial difficulty with the \texttt{pick-place-v2} and \texttt{bin-picking-v2} environments, which will affect bi-hemispheric results for these tasks.

\subsection{Main experiments}
\label{supp:main_exps}

\begin{figure}
  \centering
  \begin{subfigure}[]{\textwidth}
    \centering
    \includegraphics[width=0.8\textwidth]{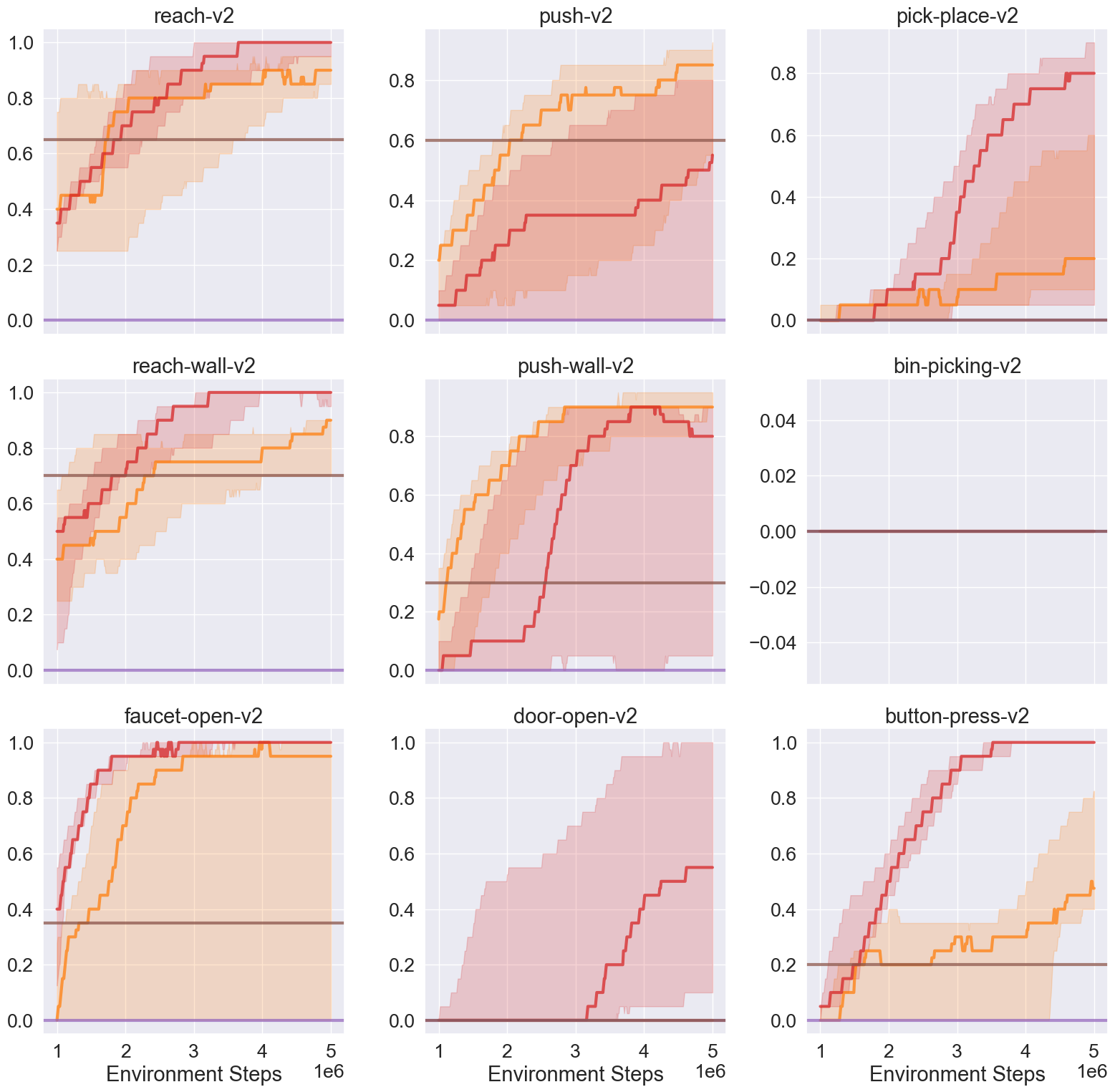}
    \caption{Success rate}
  \end{subfigure}
  \begin{subfigure}[]{\textwidth}
      \centering
      \includegraphics[width=0.8\textwidth]{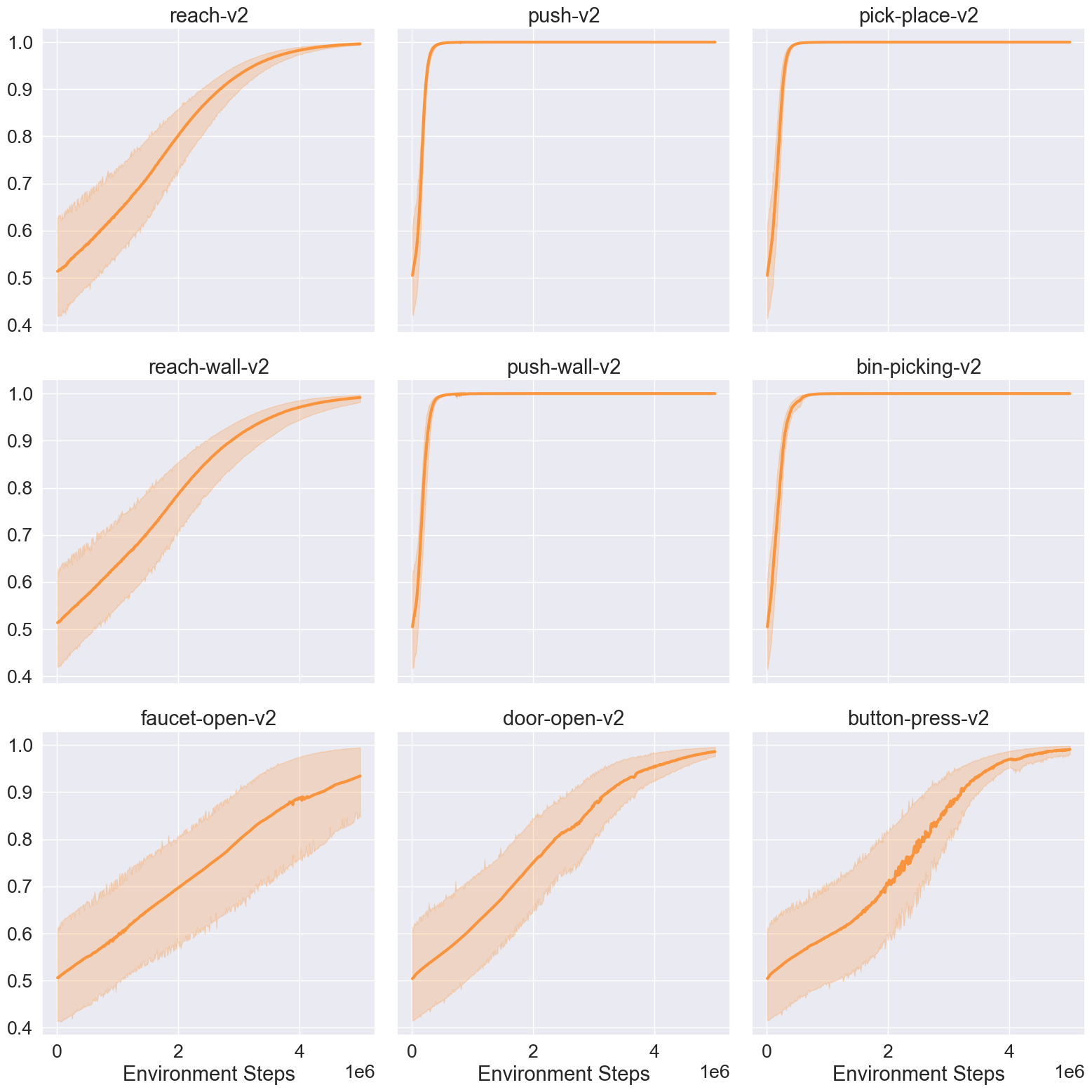}
      \caption{Left hemisphere gating values}
  \end{subfigure}
  \includegraphics[width=0.4\textwidth]{full_legend}
  \caption{Bi-hemispheric training plots}
  \label{fig:additional_bihem_results}
\end{figure}

\begin{figure}
  \centering
  \begin{subfigure}[]{\textwidth}
    \centering
    \includegraphics[width=0.8\textwidth]{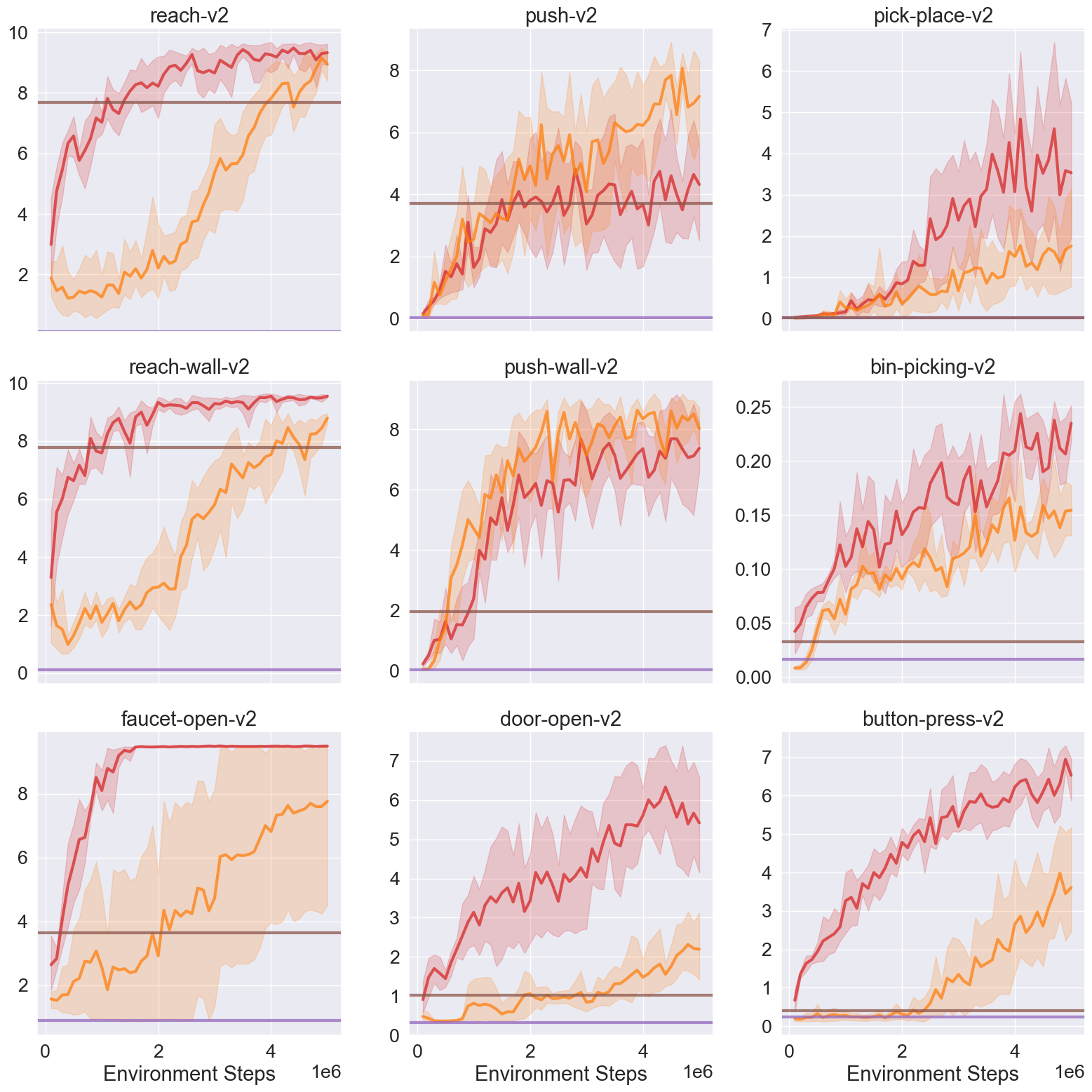}
    \caption{Average rewards}
  \end{subfigure}
  \begin{subfigure}[]{\textwidth}
      \centering
      \includegraphics[width=0.8\textwidth]{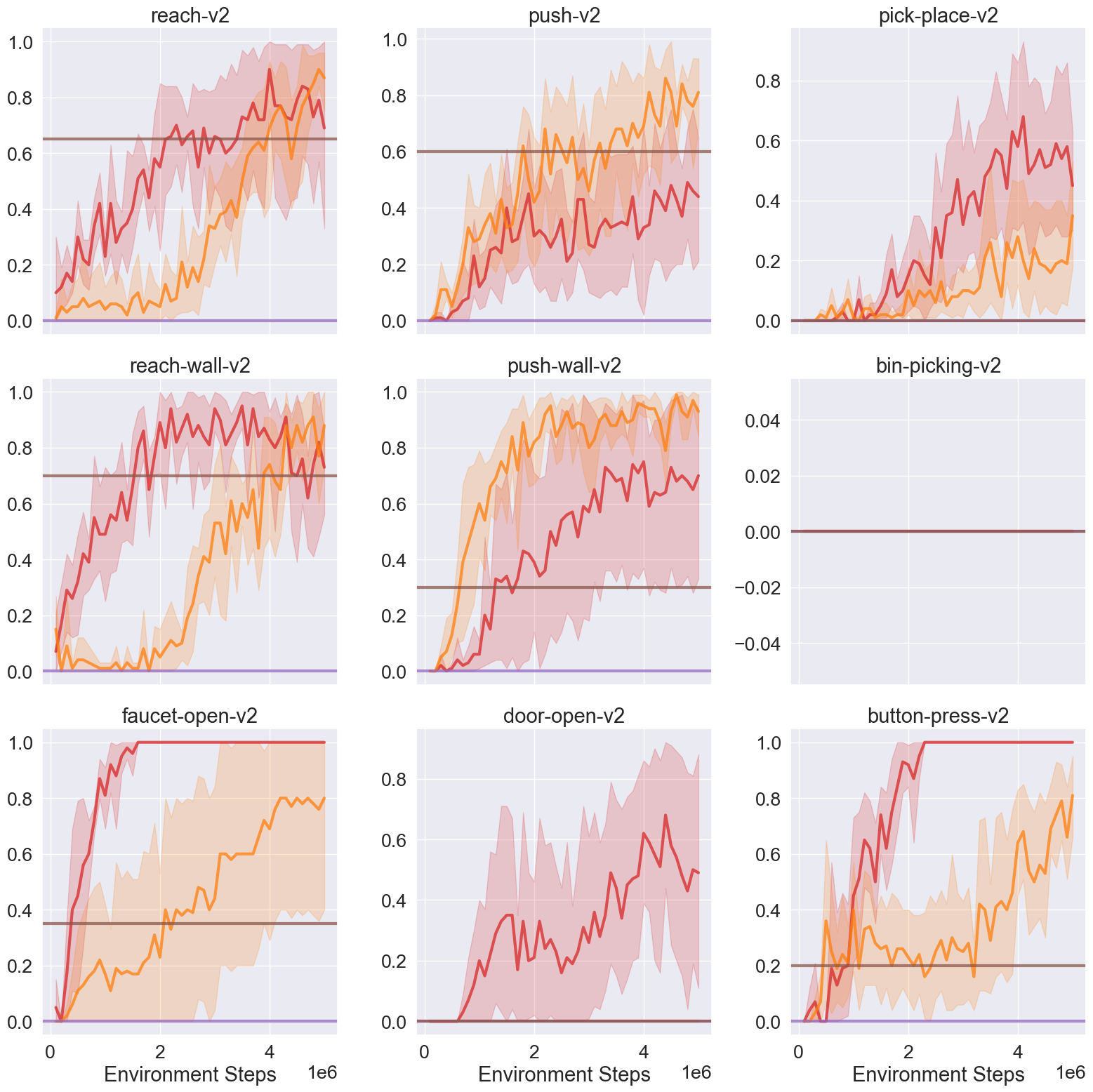}
      \caption{Success rate}
  \end{subfigure}
  \includegraphics[width=0.4\textwidth]{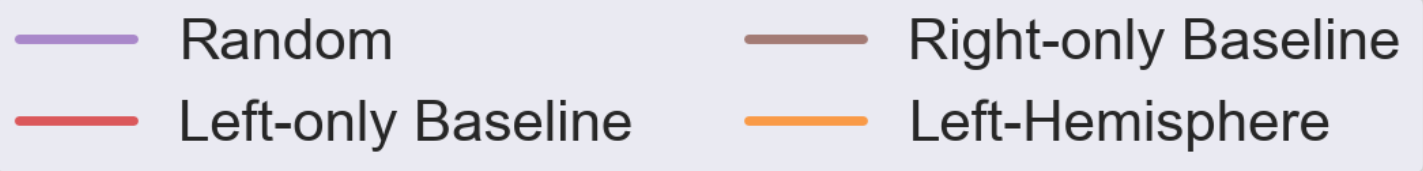}
  \caption{Left hemisphere vs baselines}
  \label{fig:left_hem}
\end{figure}

Figure~\ref{fig:additional_bihem_results} shows the success rate for the bi-hemispheric and left-only baseline agents and gating values during training.
Figure~\ref{fig:left_hem} shows the rewards and success rate for the left-hemisphere (evaluated alone) and baseline agents. 

\clearpage

\bibliographystyle{splncs04}
\bibliography{references}